\RequirePackage{fix-cm}
\documentclass[twocolumn]{svjour3}          
\smartqed  
\usepackage{color}
\usepackage{graphicx}
\usepackage{fancybox}
\usepackage{algorithmic}
\usepackage{algorithm}
\usepackage{amsmath}
\usepackage{comment}
\usepackage{bm}
\usepackage{ctable}
\usepackage{version}	
\usepackage{amsmath,amssymb,amsfonts}

\begin{document}

\title{%
A Study on Constraint Extraction and Exception Exclusion 
in Care Worker Scheduling%
\thanks{%
This work was supported by 
JST A-STEP
JPMJTM22ES.
}
}

\author{Koki Suenaga \and Tomohiro Furuta \and Satoshi Ono}

\institute{
  Koki Suenaga \at
  Information Science and Biomedical Engineering Program, Department of Engineering, Graduate School of Science and Engineering, Kagoshima University, Kagoshima, Japan \\
  \email{k0517050@kadai.jp}
  \and
  Tomohiro Furuta \at
  Information Science and Biomedical Engineering Program, Department of Engineering, Graduate School of Science and Engineering, Kagoshima University, Kagoshima, Japan \\
  \email{k6294062@kadai.jp}
  \and
  Satoshi Ono \at
  Information Science and Biomedical Engineering Program, Department of Engineering, Graduate School of Science and Engineering, Kagoshima University, Kagoshima, Japan \\
  \email{ono@ibe.kagoshima-u.ac.jp}
}

\date{Received: date / Accepted: date}

\maketitle

\begin{abstract}
Technologies for automatically generating work schedules have been extensively studied; 
however, in long-term care facilities, conditions vary between facilities, 
making it essential
to interview the managers who create shift schedules to design facility-specific constraint conditions.
The proposed method utilizes constraint templates to extract combinations of various components, 
such as shift patterns for consecutive days or staff combinations.
The templates can extract a variety of constraints by changing the number of days and 
the number of staff members to focus on and changing the extraction focus to patterns or frequency.
In addition, unlike existing constraint extraction techniques, this
study incorporates mechanisms to exclude exceptional constraints. 
The extracted constraints can be employed by a
constraint programming solver to create care worker schedules.
Experiments demonstrated that our proposed method successfully created schedules that satisfied all hard constraints and 
reduced the number of violations for soft constraints by circumventing the extraction of exceptional constraints.
\end{abstract}

\section{Introduction}
The nursing care industry has recently been facing a severe labor
shortage, and reducing staff turnover has become a critical issue
alongside efforts to recruit new personnel.
Creating work shifts that satisfy 
staffing demands is an essential measure for attracting personnel
and 
retaining employees~\cite{Agboola2016,Kurokawa2017_e}.
Care worker scheduling
resembles nurse scheduling
problems (NSP)~\cite{Ceschia2023};
each must consider
primary constraints, such as the number of
staff needed, 
each staff member's workdays, work hours, and leave requests,
and additional factors such as
interpersonal relationships among the staff.
Creating a care worker schedule typically requires an entire workday.

Technologies for automatically generating work schedules have
been extensively studied~\cite{Ceschia2023};
however, in long-term care facilities,
conditions differ significantly across facilities, and 
it is necessary to define constraints for each.
Consequently,
many interviews are required to obtain sufficient constraints
to create a schedule automatically.
Such interviews are a heavy burden for the facilities, which
has hindered the introduction of the work
schedule generation system, and very few facilities have introduced
such a system ~\cite{van2013}.
Furthermore,
compared to nurse scheduling, care worker scheduling 
is subject to various constraints,
such as the number of available shifts varying significantly 
from person to person, 
the qualification status of staff varying,
and many part-time staff working short hours.

This study proposes a method for extracting constraints 
from past schedules and generating care worker schedules using 
the extracted constraints. 
The proposed method employs constraint templates to extract 
combinations of various components, 
such as shift patterns for consecutive days or staff combinations.
In addition, unlike existing constraint extraction techniques, 
this study incorporates mechanisms to exclude exceptional constraints. 
That is, in past schedules, if there were days or a staff member with
many leave requests, 
shifts assigned to them to compensate for staff shortages---despite
being undesirable---are excluded from constraint analysis.

\begin{figure}[t]
  \centering
  \hspace*{-10mm}
  \includegraphics[width=89mm]{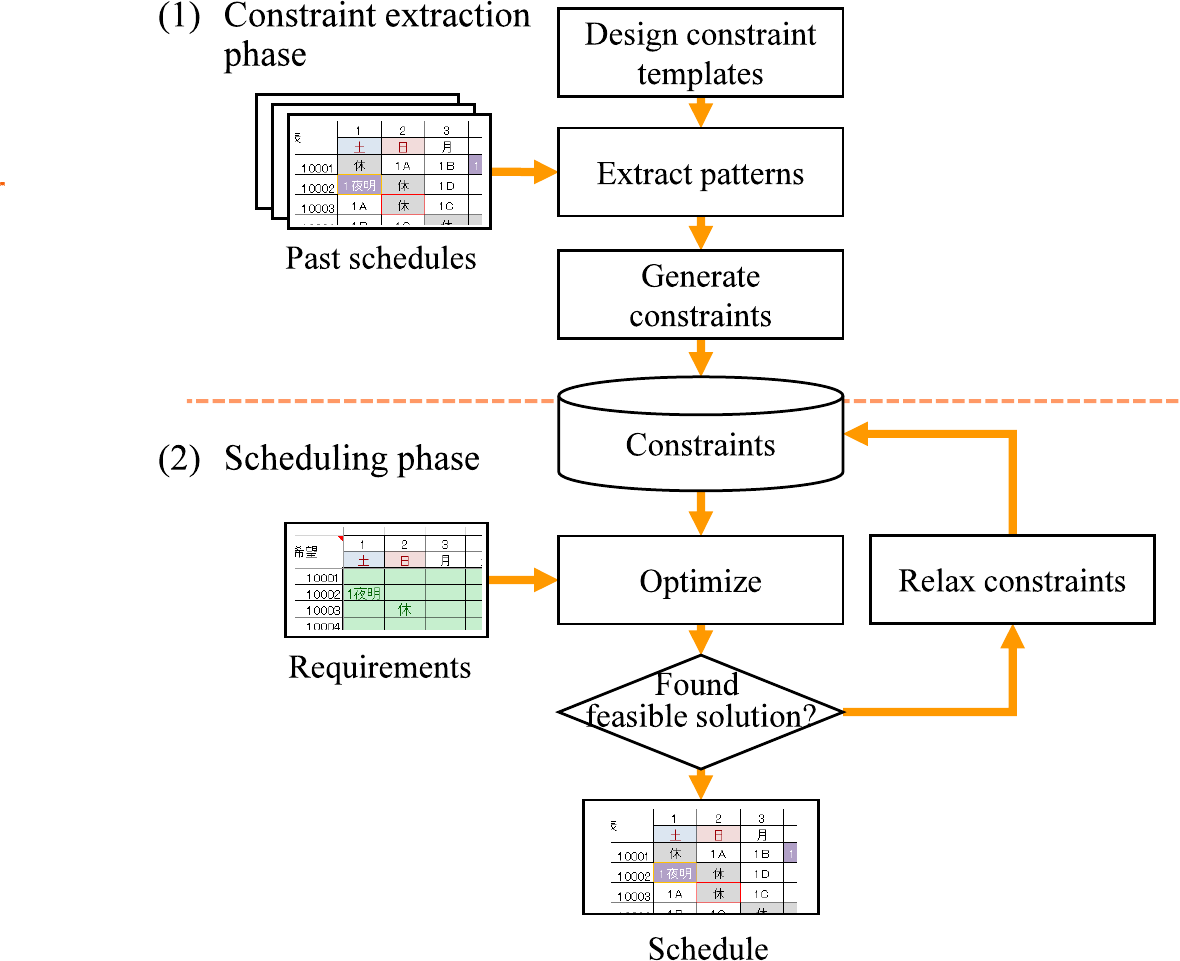}
  \caption{Overview of the proposed method, which consists of constraint extraction and scheduling phases.}
  \label{fig:overview}
\end{figure}

\section{Related work}
\subsection{Care worker scheduling}

Care worker scheduling problems are typically categorized into home-
and facility-based types~\cite{Agboola2016}.
The home-based type involves problems
of determining dates and times for service and the period of service
provided to a person in need of home care services.
The facility-based type involves problems of determining shift assignments that
respect
staff leave requests and other preferences while
promoting
fairness in the working environment.
Compared with NSPs,
facility-type care worker scheduling problems are
often over-constrained, making it difficult to obtain feasible solutions 
due to
budget
limitations and staffing shortages.
Therefore, this study focuses on care worker scheduling problems in facilities.
Kurokawa et al. solved the problems of maximizing overall sales by
accumulating services by packing as many as possible into the hours when helpers are on duty while maintaining
fairness in the working environment.
They used genetic algorithms,
differential evolution, tabu search, and quasi-annealing methods and
compared the results~\cite{Kurokawa2017_e}.
Sakaguchi et al. proposed a schedule planning and caregiver assignment
method that equalizes the workload of each care
worker~\cite{Sakaguchi2007_e}.

\subsection{Constraint extraction}
Research on the automatic design of constraints from past
solutions has recently emerged.
Constraints are essential when formulating real-world problems.
Fajemisin et al.
formalized the process of learning constraints from data,
and reviewed recent literature on constraint
learning~\cite{Fajemisin2024}.

Studies have also begun to be conducted on NSPs~\cite{kumar2019automating}.
Paramonov et al. proposed a method for extracting constraints from
two-dimensional tabular data using constraint
templates~\cite{paramonov2017tacle}.
Their method aggregates data within specific blocks and extracts
element counts and combinations according to specified criteria.
Kumar et al. proposed a method for extracting constraints from
multidimensional design variables using tensor
expressions~\cite{kumar2019automating}.
This method uses aggregate functions to extract constraints that
include numerical terms.
Ben et al. focused on NSP's dynamic characteristics, including
unpredictable and unforeseen events such as accidents
and nurses' sick leave.
They proposed a method for automatically and
implicitly learning NSP's constraints and preferences from available
historical data without prior knowledge~\cite{Ben2021}.

Although these previous studies have examined methods for extracting
diverse constraints, the problem of identifying undesirable patterns
within the extracted constraints remains underexplored.

\begin{table*}[t]
    \centering
    \caption{Constraint templates.}
    \label{tbl:constraint_template}
    {\fontsize{8pt}{9pt}\selectfont
      \begin{tabular}{@{}c@{~}|@{~}l@{~}|@{~}l@{~}|@{~}l@{~}|@{~}l@{~}|@{~}l@{~}|@{~}l@{}}
    \hline 
    ID & Template & Duration & Staff & Generality &Type & Reference \\
    \hline 
    $T_1$ & a shift pattern for $n$ consecutive days & $n$ days & each staff & specific & pattern & - \\
    $T_2$ & a shift pattern for $n$ consecutive days & $n$ days & all staff & general  & pattern & - \\
    $T_3$ & number of times a shift appeared         & a month  & each staff & specific &   count & - \\
    $T_4$ & number of persons for each shift         & a day    & all staff  & specific &   count & day of week \\
    \hline
    \end{tabular}}
~\\ ~\\
    \centering
    \caption{Relationship between constraints
      and constraint templates.}
    \label{tbl:relation_template_and_constraint}
      {\fontsize{8pt}{9pt}\selectfont
    \begin{tabular}{@{}c@{~}|@{~}c@{~}|@{~}l@{~}|@{~}c@{}}
    \hline 
      Type & ID & Constraint obtained by interview to managers & Template\\
    \hline
    Hard
    & $H_1$ & Keep the number of working days within the specified range. & $T_3$ \\
    & $H_2$ & Assign only shifts the staff is able to work. & $T_3$ \\
    & $H_3$ & Keep the working hours within the specified limit. & $T_3$ \\  
    & $H_4$ & Assign the required number of staff for each shift. & $T_4$ \\
    & $H_5$ & Assign $\rm N_{i}$ and $\rm N_{o}$ to two consecutive days. & $T_2$ \\
    \hline
    Soft
    & $S_1$ & Assign a day off the day after the $\rm N_{o}$. & $T_1$ \\
    & $S_2$ & Assign a shift sequence of $\rm N_{o}$, D, and holiday if a holiday cannot be assigned to the day after $\rm N_{o}$. & $T_1$ \\
    & $S_3$ & Prohibit a sequence of $\rm N_{o}$, a day off, and $\rm N_{i}$. & $T_1$ \\
    & $S_4$ & Equalize the number of night shift assignments. & $T_3$ \\
    & $S_5$ & Prohibit more than four consecutive day shifts. & $T_1$ \\
    & $S_6$ & Assign the requested days off or shifts to staff according to their requests. & -- \\
    \hline
    \end{tabular}}
\end{table*}

\section{The proposed method}

\subsection{Key idea}
This paper proposes a method to extract constraints from past
schedules of a long-term care facility for elderly individuals using
constraint templates and to generate a monthly daily care worker
schedule based on the extracted constraints.
Following are the key ideas of the proposed method.

\hspace*{-3ex}
{\bf Idea 1: Extracting constraints from past work schedules.}
The proposed method extracts constraints from past work schedules using
constraint templates.
The templates can extract a variety of constraints by changing the
number of days and the number of staff to focus on and changing the
extraction target to patterns or frequency.

\hspace*{-3ex}
{\bf Idea 2: Incorporating staffing margin to exclude exceptional constraints.}
If exceptional work patterns caused by understaffing, as described
above, are extracted, patterns that must be avoided, such as long
consecutive shifts, will be allowed when creating a new work schedule.
We exclude such patterns by introducing
staffing margin and flexibility, which express the ratio of available workers
relative to the number of necessary workers and the ratio of the number of requested
leave days relative to the number of the leave days, respectively.

\hspace*{-3ex}
{\bf Idea 3: Employing constraint programming solver with gradual constraint relaxation.}
Any solver can use the extracted constraints, such as linear
programming, constraint programming, and meta-heuristics.
This study adopts a constraint programming solver to
validate the extracted constraints' effectiveness.
The solver disallows assignments outside the extracted patterns, yet
some valid assignments may not have appeared in past schedules.
Therefore, when no feasible solution is found, the proposed method
gradually relaxes some of the hard constraints into soft constraints
to obtain a feasible solution.

\subsection{Overview of the proposed method}

Fig.~\ref{fig:overview} shows the overall structure of the proposed method, which
includes 
a constraint extraction phase and a scheduling phase. 
In the constraint extraction phase, constraints are extracted from
past work schedules using constraint templates, which are work
patterns or the number of occurrences of patterns in a range of
interest from past work schedules.

The scheduling phase generates a new schedule using the extracted
constraints.
In the scheduling phase, 
various methods, such as constraint
satisfaction solvers and meta-heuristics, can be used in the proposed
method.

\subsection{Formulation of scheduling}

In the scheduling phase, the solver decides a set $\mathcal{X}$ of binary variables that indicate,
for each day $d$, staff member $w$, and shift $s$, whether $w$ takes $s$ on $d$ (1) or not (0).
\begin{equation}
\mathcal{X} = \left\{
                x_{(d, w, s)}
              \right\}_{d \in \mathcal{D}, w \in \mathcal{W}, s \in \mathcal{S}}
~~~
(x_{d, w, s} \in \{0, 1\})            
\end{equation}
where $\mathcal{D}$ denote the set of days to be scheduled; our
proposed method builds monthly rosters, so $|\mathcal{D}| \in \{28,
\ldots, 31\}$.
$\mathcal{W}$ and $\mathcal{S}$ denote the staff set and the set of
shift types, respectively.

The staff rostering problem can be formulated as follows:
\begin{align}
{\rm minimize}~~~    & \sum_{i \in C_{\rm soft}} V_i(\bm{x}) 
\\
{\rm subject~ to}~~ & V_j (\bm{x}) = 0 ~~ (\forall j \in C_{\rm Hard})
\end{align}
where $V_i(\bm{x}) = 1$ if $\bm{x}$ violates constraint $i$ and $0$ otherwise;
$C_{\rm Soft}$ and $C_{\rm Hard}$ collect 
soft and hard
constraints, respectively.
Thus, the solution satisfies all 
hard
constraints while reducing
violations of 
soft
constraints.
For instance, a constraint to ensure that staff $w$ takes exactly one
shift on day $d$, can be described as follows:
\begin{equation}
{\rm subject~ to}~~
\sum_{s \in \mathcal{S}} x_{d, w, s} = 1.
\end{equation}
Likewise, a constraint to satisfy demand for shift $s$ on day $d$ 
is 
\begin{equation}
{\rm subject~ to}~~
\sum_{w \in \mathcal{W}} x_{d, w, s} \geq r_{d, s},
\end{equation}
that is, at least $r_{d, s}$ staff for shift $s$.

\subsection{Templates for constraint extraction}

The specified shift combinations are retrieved from past 
schedules using constraint templates in the constraint extraction
stage.
These templates  
specify the shift and staff combinations to be extracted, and 
those that appear are considered assignable combinations.
Table~\ref{tbl:constraint_template}
shows the constraint templates applied in this method.
The constraint template
comprises
the period, number of persons,
generality,
type,
and reference information.
The number of days determines the length of time for extracting the
pattern and the number of times. 
The number of persons determines how many patterns are extracted.
Generality specifies whether the focus is on particular elements
(e.g., members) or on all elements,
while type
determines whether the pattern is extracted or how many 
times it is measured.  
Reference determines which data are referred to when extracting
elements other than past schedules, such as days of the week and
requested holidays.

Constraint templates $T_1$ and $T_2$ extract shift patterns a
staff member has worked consecutively.
The extracted constraints are formalized
and passed to the solver in the scheduling phase.
We specifically extract the assignable shifts obtained using $T_1$ for
each staff member individually, and extract the assignable shifts for
all staff members using $T_2$.
Constraint template $T_3$ extracts the number of days per month that
each staff member is available to work.
Constraint template $T_4$ extracts the number of shifts required for
each day of the week.
The above template is necessary because the number of required workers 
changes for each day of the week.

Table \label{tbl:relation_template_and_constraint} shows the
relationship between the constraint templates and general constraints
obtained by interviews to a facility manager. 
Constraints are generally categorized as hard constraints that must be
satisfied and soft constraints that should be satisfied as far as possible.
Hard constraints $H_1$ and $H_3$ limit monthly working days and
working hours, and $T_3$ derives these limits from past schedules.
Likewise, $H_2$ guarantees that each staff member receives only shifts
they can cover, and $T_3$ extracts this constraint as well.
$H_5$ addresses night work that spans two dates; $N_i$ and $N_o$
denote the shifts on the start and end days, respectively.
Constraints of this type follow from template $T_2$. 
In the facility we interviewed, staff take the day after a night shift
off; however, when applying our method to other facilities that assign
consecutive nights, $T_2$ still captures the corresponding
constraints.

Preferences about night work are modeled as soft constraints, and
template $T_1$ extracts them automatically.
For example, the interviewed facility requires a day off after a night
shift ($S_1$).
When this proves difficult, it assigns a day shift on the following
day and then grants a day off after that day shift ($S_2$).
It also permits a night-off-night pattern ($S_3$).
$T_1$ derives these rules from historical rosters.
In addition, $T_1$ extracts a constraint that limits consecutive day
shifts to at most three days ($S_5$), while $T_3$ balances night-shift
assignments across staff ($S_4$) by using shift-count statistics.

As described above, most constraints fit one of four templates;
accordingly, we include $S_6$ by a default constraint because $S_6$ is
common across facilities.

\begin{algorithm}[t]
  \caption{Constraint extraction using templates~\protect\footnotemark[1]}
  \label{fig:alg_extract}
{\fontsize{8pt}{10pt}\selectfont
\begin{algorithmic}[1]
  \REQUIRE Staff set $E$, shift set $S$, historical rosters $P$, 
           shift and vacation requests $Q$
           integers $n^{\min}_d \le n^{\max}_d$, thresholds $\tau_u,\tau_c,\tau_f$
  \ENSURE Constraint set $C$
  \STATE {\it \# Build templates}
  \STATE $\mathcal{T} \gets \emptyset$
  \FOR{$n' \gets n^{\min}_d$ \TO $n^{\max}_d$}
    \STATE $\mathcal{T} \gets \mathcal{T} \cup 
            \{ (\delta{=}n', \sigma{=}\textsc{1}, \Phi{=}\textsc{Pattern}, \gamma{=}\textsc{specific}  ) \}$ 
           \textit{\# ~$T_1$}
    \STATE $\mathcal{T} \gets \mathcal{T} \cup 
            \{ (\delta{=}n', \sigma{=}\textsc{1}, \Phi{=}\textsc{Pattern}, \gamma{=}\textsc{general} ) \}$ 
           \textit{\# ~$T_2$}
  \ENDFOR
  \STATE $\mathcal{T} \gets \mathcal{T} \cup 
          \{ (\delta{=}\mathrm{MONTH}, \sigma{=}\textsc{1}, \Phi{=}\textsc{Count}(\textsc{ANY}, \textsc{UNIT}), $
  \item[] $\gamma{=}\textsc{specific}  ) \}$ 
         \textit{\# ~$T_3$}
  \FOR{\textbf{each } $s_0 \in S$}
    \STATE $\mathcal{T} \gets \mathcal{T} \cup 
          \{ (\delta=1, \sigma{=}\textsc{ALL}, \Phi{=}\textsc{Count}(s_0, \textsc{WEEKDAY}), $
    \item[] $\gamma{=}\textsc{general} ) \}$ 
          \textit{\# ~$T_4$}
  \ENDFOR
  \STATE {\it \# Constraint extraction main loop}
  \FOR{\textbf{each } $t=(\delta, \sigma, \Phi, \gamma) \in \mathcal{T}$}
    \STATE $\Sigma_t \gets \emptyset$ ;~ $n_m \gets 0$ 
    \FOR{\textbf{each } month $m \in P$}
      \STATE $n_m \gets n_m + 1$
      \STATE $D_m \gets \emptyset$ 
      \STATE {\it \# Exclude staff whose flexibility is less than $\tau_f$}
      \STATE $E_{ok} \gets \textsc{Eligibles}(E, m, Q, \tau_f)$
      \STATE $n_{\mathrm{eff}} \gets (\delta = \mathrm{MONTH}) ~?~ \text{last\_day}(m) ~:~ \delta  $
      \FOR{$D \gets 1$ \TO $\text{last\_day}(m) - n_{\mathrm{eff}} + 1$}
        \STATE $W \gets [D,\, D+n_{\mathrm{eff}}-1]$
        \STATE {\it \# Extraction considering staffing margin} 
        \IF{$u_d \geq \tau_u ~\forall d \in W$}
          \STATE $D_m \gets D_m \uplus \mathbf{Collect}(t, E_{ok}, S, m, W)$
        \ENDIF
      \ENDFOR
      \STATE $\Sigma_t \gets \textbf{ReduceMonth}(D_m,\Sigma_t)$ 
    \ENDFOR
    \STATE $C \gets C \cup \textbf{ReduceFinal}(t, \Phi, \Sigma_t, n_m,\tau_c)$
  \ENDFOR
  \STATE \textbf{return } $C$
\end{algorithmic}
}
\end{algorithm}
\footnotetext[1]{
  \textit{Notation.} For a multiset $M$, 
  $\mu_M(\pi)$ denotes the multiplicity (count) of pattern $\pi$ in $M$; $\uplus$ is multiset addition; 
  $\mathrm{sup}p(M) = \{x \in U | \mu_M(x) > 0 \}$ .
}

\begin{algorithm}[t]
  \caption{%
    Collect$(t=(\delta, \sigma, \Phi, \gamma),E,S,m,W)$ 
  }
  \label{fig:alg_extract_collect}
{\fontsize{8pt}{10pt} \selectfont
\begin{algorithmic}[1]
  \STATE $\Delta \gets \emptyset$
  \IF{$\Phi = \textsc{Pattern}$ \AND $\sigma=\textsc{1}$}
    \FOR{\textbf{each } $e \in E$}
       \STATE $\pi \gets (\textsc{Roster}(e,m,d'))_{d' \in W}$
        \STATE $\kappa \gets \textbf{ComposeKey}(\gamma, e, m, W, \pi)$ 
        \STATE $\Delta \gets \Delta \uplus \{\,(\kappa, 1)\,\}$
    \ENDFOR
  \ELSIF{$\Phi=\textsc{Count}(\theta,\nu)$}
    \FOR{\textbf{each } $d' \in W$}

      \IF{$\sigma=\textsc{1}$}  
        \FOR{\textbf{each } $e \in E$}
          \STATE $s \gets \textsc{Roster}(e, m, d')$
          \STATE $\kappa \gets \textbf{ComposeKey}(\gamma, e, m, [d', d'], s)$  
          \STATE $\Delta \gets \Delta \uplus \{\,(\kappa, 1)\,\}$
        \ENDFOR

    \ELSIF {$\sigma=\mathbf{ALL}$}   

      \STATE $w \gets \textsc{Weekday}(m,d')$
      \STATE $\textit{Tgts} \gets (\theta=\textsc{ANY}) ~?~ S  ~:~ \{\theta\}$
      \FOR{\textbf{each } $t_{d'} \in \textit{Tgts}$}
        \STATE $h \gets \bigl|\{\, e\in E \mid \textsc{Roster}(e,m,d')=t_{d'} \,\}\bigr|$  
        \STATE $u \gets \textsc{Normalize}(\nu, w, h, m)$
        \STATE $\kappa \gets \textsc{ComposeKey}(\gamma,\ \textit{nil},\ m,\ [d',d'],\ (w,t_{d'}))$
        \STATE $\Delta \gets \Delta \uplus \{\,(\kappa, u)\,\}$
      \ENDFOR
    \ENDIF
  \ENDFOR
  \ENDIF
  \STATE \textbf{return } $\Delta$
\end{algorithmic}
}
\end{algorithm}

\begin{algorithm}[t]
  \caption{ReduceMonth$(D_m,\Sigma_t)$
    }
  \label{fig:alg_reduce_month}
{\fontsize{8pt}{10pt}\selectfont
\begin{algorithmic}[1]
  \STATE $\Sigma'_t \gets \Sigma_t$
  \FOR{\textbf{each } $\kappa \in \{\,\kappa' \mid \exists u:\ (\kappa',u)\in \mathrm{supp}(D_m)\,\}$}
    \STATE $U_\kappa \gets 0$
    \FOR{\textbf{each } $u \text{ with } (\kappa,u)\in \mathrm{supp}(D_m)$}
      \STATE $U_\kappa \gets U_\kappa + \mu_{D_m}(\kappa,u)\cdot u$ 
    \ENDFOR
    \STATE $\Sigma'_t \gets \Sigma'_t \uplus \{\, (\kappa, U_\kappa) \,\}$
  \ENDFOR
  \STATE \textbf{return } $\Sigma'_t$
\end{algorithmic}
}
\end{algorithm}
\begin{algorithm}[t]
  \caption{ReduceFinal$(t, \Phi, \Sigma_t, n_m, \tau_c)$}
  \label{fig:alg_extract_reduce}
{\fontsize{8pt}{9pt}\selectfont
\begin{algorithmic}[1]
  \STATE $C_T \gets \emptyset$
  \IF{$\Phi = \textsc{Pattern}$} 
    \FOR{\textbf{each } $\kappa$ \textbf{ with } $(\kappa,v)\in \mathrm{supp}(\Sigma_t)$}
      \STATE $V_\kappa \gets \{\, v \mid (\kappa,v)\in \mathrm{supp}(\Sigma_t)\,\}$
      \STATE $\bar{v}_\kappa \gets \sum_v \mu_{\Sigma_t}(\kappa, v) \cdot v  / n_m$
      \IF{$\bar{v}_\kappa \ge \tau_c$} \STATE $C_T \gets C_T \cup \{\kappa\}$ \ENDIF
    \ENDFOR
    \STATE \textbf{return } $C_T$
  \ELSIF{$\Phi = \textsc{Count}(\cdot, \cdot)$} 
    \FOR{\textbf{each } $\kappa$ \textbf{ with } $(\kappa,v)\in \mathrm{supp}(\Sigma_t)$}
      \STATE $V_\kappa \gets \{\, v \mid (\kappa,v)\in \mathrm{supp}(\Sigma_t)\,\}$
      \STATE $\underline{b} \gets \lfloor \min V_\kappa \rfloor$;\quad $\overline{b} \gets \lceil \max V_\kappa \rceil$
      \STATE $C_T \gets C_T \cup \{\, (\kappa,\ \underline{b},\ \overline{b}) \,\}$
    \ENDFOR
    \STATE \textbf{return } $C_T$
  \ENDIF
\end{algorithmic}
}
\end{algorithm}

\subsection{Extraction considering staffing margin and flexibility}

The proposed method incorporates a staffing margin to exclude the
workdays with staff shortages from constraint extraction.
The staffing margin $u_d$ for a given workday $d$ is the value obtained by
dividing the number of staff available for work $a_d$ by the required
number of staff $r_d$, i.e., $u_d = \frac{a_d}{r_d}$.
Note that $u_d$ can be calculated from the results of staff members'
vacation requests.
If $u_d$ is less than the threshold value $\tau_u$, the
constraint is not extracted for day $d$.
When applying a constraint template that spans multiple days,
the minimum staffing margin among those days is used.

The proposed method also considers a staff member that requests many
leave days as an exception, because too many requests makes it more
difficult to create a reasonable schedule.
To extract such staff, the proposed method calculates flexibility $u_f$.

The flexibility $u_f$ for a staff member is calculated as the ratio
$1 - N_d^{(r)} / N_d^{(a)}$, 
where $N_d^{(r)}$ signifies the number of
requested leave days and $N_d^{(a)}$ represents the number of assigned
days.
Staff members whose flexibility
is less than
the threshold $\tau_f$ are
excluded from constraint extraction for that month.

In addition to excluding exceptions using the aforementioned $u_d$ and
$u_f$, the proposed method also omits shift patterns with low frequency.
That is, if the number of occurrences of an extracted pattern 
is less than the threshold value
$\tau_c$, it is not extracted as a constraint.

Algorithms~\ref{fig:alg_extract}, \ref{fig:alg_extract_collect},
\ref{fig:alg_reduce_month}, and \ref{fig:alg_extract_reduce} show the
algorithm for extracting constraints using constraint templates $T_1$
through $T_4$ while considering staffing margin.
The main loop, shown in
Algorithm~\ref{fig:alg_extract}, runs per
constraint template and per month.
Each constraint template includes duration $\delta$, target staff
$\sigma$, generality $\Phi$, and type $\gamma$.
Using the template-defined sliding window $W$, our method scans
historical rosters.
It invokes procedure Collect shown in
Algorithm~\ref{fig:alg_extract_collect} to list patterns and their counts,
then invoke procedure ReduceMonth shown in
Algorithm~\ref{fig:alg_extract_reduce} for monthly aggregation.
Procedure Collect scans the past scheduling table using window $W$, and
generates a multiset $\Delta$ of $(\kappa, u)$ pairs.
Its subprocedure ComposeKey generates the aggregation key $\kappa$
while encoding the generality (e.g., distinguish $T_1$ and $T_2$).
After all months are processed, it run procedure ReduceFinal shown in
Algorithm~\ref{fig:alg_extract_reduce} to aggregate patterns and produce
constraint instances.
Procedure ReduceFinal aggregates the monthly summaries $\{\Sigma_m\}$
by key across months, calls Normalize to compute the
template-specified metric, and converts the result into the final
constraint instances to return.
Subprocedure Normalize converts the daily staff count $h$ to a
cross-month scale; it returns $h$ for COUNT type template with UNIT
option, and $h$ divided by its number of occurrence days for a
template with WEEKDAY option.

\subsection{Scheduling with gradual constraint relaxation}
\label{ssec:relax}

The scheduling phase generates shift plans based on extracted
constraints.
Constraints specifying
exact 
numbers of working days or staff members, 
such as $H_1$ and $H_4$, 
are
implemented in 
as soft constraints for the exact target and 
as hard
constraints that allow minor deviations, enabling practical and
flexible assignments.

When no feasible solution is found, the proposed method gradually
relaxes 
hard
constraints into soft ones and repeatedly apply the
solver.
Constraints extracted from Template $T_2$ are initially treated as
hard constraints.
However, since they prohibit all shift patterns that did not appear in
past schedules, they can be overly restrictive.
Moreover, they may conflict with other hard constraints and prevent
scheduling generation.
Thus, when infeasibility occurs, the proposed method begins by relaxing
longer constraints from $T_2$ and reattempt solving.
The similar staff members were identified using attributes that follow from
contract type, such as feasible shift types and the number of working days.

In addition to the constraints extracted from the past schedules,
the following constraints were manually added: constraints that were
easily clarified during
interviews and constraints
related to newly hired staff, which were difficult to extract from past
work schedules.
\begin{itemize}

\item 
  Constraints for new staff members not included in past schedules
  used for extraction were defined by mirroring those
  of similarly positioned staff.

\item 
  Constraints $S_6$, which fulfills leave requests and shift preferences, was initially added
  as hard constraints.
If the solution remains infeasible, the proposed method proceeds to
incrementally remove shift requests by staff members to further relax
the constraints.

\end{itemize}

\begin{algorithm}[t]
  \caption{Scheduling with constraint relaxation}
  \label{fig:}
{\fontsize{8pt}{10pt}\selectfont
\begin{algorithmic}[1]
\REQUIRE 
  constraints $C_{Hard}$ and $C_{Soft}$ ; 
  constraint template $T_2$
\STATE $(ok,Plan) \gets \textsc{Solve}(C_{Hard},C_{Soft})$; 
\STATE {\textbf{if } {$ok$} \textbf{then return } $Plan$}
\WHILE{not $ok$}
  \STATE $C_r \gets$ longest subset of $(T_2 \cap C_{Hard})$
  \STATE \textbf{if } $C_r = \emptyset$ \textbf{ then break} 
  \STATE $C_{Hard} \gets C_{Hard} \setminus C_r$;\; $C_{Soft} \gets C_{Soft} \cup C_r$
  \STATE $(ok,Plan) \gets \textsc{Solve}(C_{Hard},C_{Soft})$; 
\ENDWHILE
\WHILE{not $ok$}
  \STATE choose a constraint instance $c_r \in C_{Soft}$ (vacation/shift request)
  \STATE \textbf{if } no request remains \textbf{ then break}
  \STATE $C_{Soft} \gets C_{Soft} \setminus c_r$
  \STATE $(ok,Plan) \gets \textsc{Solve}(C_{Hard},C_{Soft})$; 
\ENDWHILE
\STATE \textbf{if } $ok$ \textbf{ then return } $Plan$ \textbf{ else } \textsc{Infeasible}
\end{algorithmic}
}
\end{algorithm}

\section{Evaluation}

\subsection{Experimental setup}

Experiments using 48 months of scheduling data at a special nursing
home for elderly individuals in Kagoshima city were conducted to
verify the proposed method's effectiveness.
This facility divides its living space into two units, U1 and U2;
For four years, we created work schedules for the facility, and the
facility manager revised them, thereby accumulating data.
The data included survey results of desired leave days from all
staff in addition to past work schedules.
Of the work schedules and leave requests obtained, 36 months were
used for constraint extraction, and the other 12 months were used
for work schedule design.

In this experiment, the proposed method abstracted all of the day
shifts in U1 and U2 into one type of shift, D, and the night shifts in
U1 and U2 were abstracted into 1N and 2N, respectively.
Thus, day shift collapses to one shift type D across U1 and U2, whereas
night shift preserves the unit label and unifies the night-start and
night end days, $\rm N_i$ and $\rm N_o$, into 1N (U1) and 2N (U2), as shown 
in Table~\ref{tbl:relation_shift}.
A, B, and C shifts denote day shifts with different start times (A:
early; B: standard, C: late); A shift also holds the day-shift leadership.
D denotes the day-care assignment for non-residential service users,
and E denotes a morning-only assignment.

The task of constructing detailed shifts including original 16 shift
types from the abstracted ones is considerably easier than generating
the abstract shifts themselves.
The practical shift can be obtained by subdividing the day shift D
into 10 variations.
This is because the constraints on these detailed shifts are relatively
weak compared to those for night shifts or leave, and can be satisfied
by adjustments within the subdivision framework, thus allowing the
process to be performed in a nearly greedy manner.
The quality of the final, detailed shift depends directly on the
quality of the initial abstract shifts; higher-quality abstract shifts
yield a high-quality final schedule.

We took this approach because 
the data for constraint extraction was insufficient for 
many constraints.
The relationship between the constraint templates and constraints
obtained by the hearing under the condition that abstracted day shifts, as mentioned
above, was summarized in
Table~\ref{tbl:relation_template_and_constraint}.
Templates $T_1$ and $T_2$ were applied to past schedules to extract 
patterns ranging from two to seven days in duration, 
i.e., $n_d^{\textit min} = 2$ and $n_d^{\textit max}=7$.
Parameters $\tau_c$, $\tau_u$, and $\tau_f$ were set to 0.15, 1.25, and
0.5, respectively%
\footnote{%
These thresholds were heuristically determined.
$\tau_c$ and $\tau_u$ were calculated based on the value near the 33rd
percentile of monthly occurrence rates across shift patterns (after
sorting in ascending order), and $(\mathrm{daily minimum} +
2)/(\mathrm{daily minimum})$ for day shifts (e.g., $10/8$),
respectively.
We set $\tau_f=0.5$ because months with leave requests on more than
half of the days showed many exceptional assignments.
}.

\begin{table}[t]
  \caption{Detailed and abstracted shifts.}
  \label{tbl:relation_shift}
  \centering
  {\fontsize{8pt}{9pt}\selectfont
  \begin{tabular}[0.5\textwidth]{cc@{~}|@{~}cc@{~}|@{~}cc}
  \hline
  original&abstract&
  original&abstract&
  original&abstract\\
  \hline
  day off & - & 1A & D & 2A & D \\
  paid off & - & 1B & D & 2B & D \\
  & & 1C & D & 2C & D \\
  & & 1D & D & 2D & D \\
  & & 1E & D & 2E & D \\
  & & 1Ni & 1N & 2Ni & 2N \\
  & & 1No & 1N & 2No & 2N \\
  \hline
  \end{tabular}}
\end{table}

\subsection{Experiment 1: extracting constraints from past schedules}

Table~\ref{tbl:num_constraints} shows the number of constraints
extracted from past schedules for 36 months.
Incorporating staffing margin and flexibility reduces constraints
obtained using $T_1$. 
Fig.~\ref{fig:ex_constraints} shows example constraints created using
$T_1$ for staff $10005$.
The values within each extracted constraint instance includes staff ID
and an available shift sequence.
For instance, the first instance indicates staff 10005 can take a day
shift after a day off, and the fourth instance describes they can take
a day off after a night shift and then work a day shift the following
day.
It demonstrated that the proposed method could generate essential
constraints, such as scheduling night shifts for two consecutive days
and providing a day off subsequent to night shifts.
Fig.~\ref{fig:ex_constraints_excluded} shows example constraints
excluded as exceptions using staffing margin and flexibility. 
The examples showed that schedules assigning day shifts immediately
following night shifts and schedules assigning a single day shift with
days off before and after were excluded as exceptions.

Fig.~\ref{fig:extracted_constraints_t2} shows the
extraction results using the constraint template $T_3$, which provides
a list of available shifts for each staff member.
The numerical values within each extracted constraint instance denote
the staff ID, shift type, the start and end work dates for the month,
and the lower and upper limits for the shift.  
Furthermore,
$T_3$
 told us that staff member 10006 is a part-time
staff member
because of
the small number of days they work and which
staff members need to reduce the number of night shifts.

Fig.~\ref{fig:extracted_constraints_t3} shows constraint examples extracted
using the constraint template $T_4$.
Each extracted constraint instance specifies the day of week, the
shift, and the number of staff needed.
These results indicate that
the number of day shift workers required varies
depending on the day of the week.
This is because day-care services are provided only on certain days of
the week at the facility, and the number of day shift workers is
reduced on weekends when day-care services are not provided.
In addition, two night shift workers are required for each unit
regardless of the day of the week.
This is because the staff in charge of the night shift from the day
before to the day of the experiment and the staff in charge of the
night shift from the day of the experiment to the next day are needed
separately.

\begin{table}[t]
  \centering
  \caption{The number of extracted constraints.}
  \label{tbl:num_constraints}
  {\fontsize{8pt}{9pt}\selectfont
  \begin{tabular}{@{}c@{~}|@{~~}r@{~~}r@{}}
    \hline
    Used  & \multicolumn{2}{c}{The number of constraints extracted} \\
    template              & w/o exception exclusion & w/t exception exclusion \\
    \hline
     $T_1$ & 5,867 & 4,953 \\
     $T_2$ & 1,845 & 1,845 \\
     $T_3$ &    96 &    96 \\
     $T_4$ &     7 &     7 \\
    \hline
  \end{tabular}
  }
\end{table}

\begin{figure}[t]
\centering
~\\
~\\
\begin{minipage}{0.228\textwidth}
{\fontsize{7pt}{7pt}\selectfont
\begin{Sbox}
\begin{minipage}[t]{0.95\textwidth}
\begin{verbatim}
(10005,  -,  D)
(10005, 1N,  -)
(10005,  D, 1N)
(10005, 1N,  -,  D)
(10005,  D, 1N, 1N,  -)
...
\end{verbatim}
\end{minipage}
\end{Sbox}
\fbox{\TheSbox}
}
\caption{Examples of extracted constraints using template $T_1$.}
\label{fig:ex_constraints}
\end{minipage}
~
\begin{minipage}{0.228\textwidth}
{\fontsize{7pt}{7pt}\selectfont
\begin{Sbox}
\begin{minipage}[t]{0.95\textwidth}
\begin{verbatim}
(10005, 1N,  D)
(10005, 1N, 1N,  D)
(10005,  -,  -,  D,  -)
(10005,  -,  D,  -,  D, 1N)
(10005,  D, 1N, 1N,  -, 1N)
...
\end{verbatim}
\end{minipage}
\end{Sbox}
\fbox{\TheSbox}
}
\caption{Examples of excluded exceptional constraints.}
\label{fig:ex_constraints_excluded}
\end{minipage}
~\\ ~\\ ~\\
\centering
\begin{minipage}[t]{0.232\textwidth}
{\scriptsize
\begin{Sbox}
\begin{minipage}{0.95\textwidth}
\begin{verbatim}
(10006,  -, 1, 31, 15, 15)
(10006,  D, 1, 31, 15, 15)
(10006, 1N, 1, 31,  0,  0)
(10006, 2N, 1, 31,  0,  0)
...
\end{verbatim}
\end{minipage}
\end{Sbox}
\fbox{\TheSbox}
}
\caption{Extracted constraints using template $T_3$: Available work days.}
\label{fig:extracted_constraints_t2}
\end{minipage} 
~
\begin{minipage}[t]{0.232\textwidth}
\centering
{\scriptsize
\begin{Sbox}
\begin{minipage}{0.95\textwidth}
\begin{verbatim}
("Sun.",  D, 6)
("Sun.", 1N, 2)
("Sun.", 2N, 2)
("Mon.",  D, 9)
("Mon.", 1N, 2)
("Mon.", 2N, 2)
...
\end{verbatim}
\end{minipage}
\end{Sbox}
\fbox{\TheSbox}
}
\caption{Extracted constraints using template $T_4$: Required staff for each shift for each day of the week.}
\label{fig:extracted_constraints_t3}
\end{minipage} 
\end{figure}

\subsection{Experiment 2: scheduling using extracted constraints}

Next, using the extracted constraints, the proposed method generated
care work schedules for 12 months.
The strength of the extracted constraints was manually established as
follows.
All constraints extracted by $T_1$ 
were regarded as soft 
constraints.
For the constraints extracted by $T_3$, the exact extracted count was
treated as a soft constraint, while the count within a $\pm 1$ range was
treated as a hard constraint.
For example, if the extracted number of times was 8, two constraints
were created: a soft constraint to satisfy the specified number of
times and a hard constraint to keep the number of times within
the range of seven to nine.
The conditions extracted in $T_4$ were also expressed as a combination of
hard and soft constraints
like those
extracted by $T_3$.
For example, if four staff members are required to shift D on a
particular
day, there is no constraint violation if four staff members are
assigned.
A soft constraint is violated if five staff members are
assigned, and a hard constraint is violated if more than six or fewer
than three staff members are assigned.

Constraints extracted using $T_2$ followed a different treatment.
The constraints were generally considered as 
hard
constraints;
however, if a feasible schedule could not be created, the
constraints were iteratively relaxed, i.e., regarded as soft
constraints, beginning with the longest ones, until the scheduling
succeeded.

This experiment used a CP-SAT solver 
Ver. 9.6.2534
in 
Google OR-Tools to directly verify the extracted
constraints' effect.
The soft constraints were incorporated by using the number of
violations as the objective function.
We set the solver's time limit to one hour.
Constraints extracted using $T_2$ were treated as 
hard
constraints,
as specified in Sec.~\ref{ssec:relax}, but progressively relaxed
starting from seven-day constraints when a feasible solution was
unavailable.
Solutions were identified directly under strict constraints in five of
the twelve months, while in three months, relaxation to a weaker form
with seven-day constraints was necessary. 
Moreover, additional relaxations to constraints of six, five, and four
days were each required for one month, two months, and one month,
respectively, whereas no greater relaxation was required during the
twelve-month experimental evaluation.

\begin{figure}[t]
  \centering
  \includegraphics[width=0.48\textwidth]{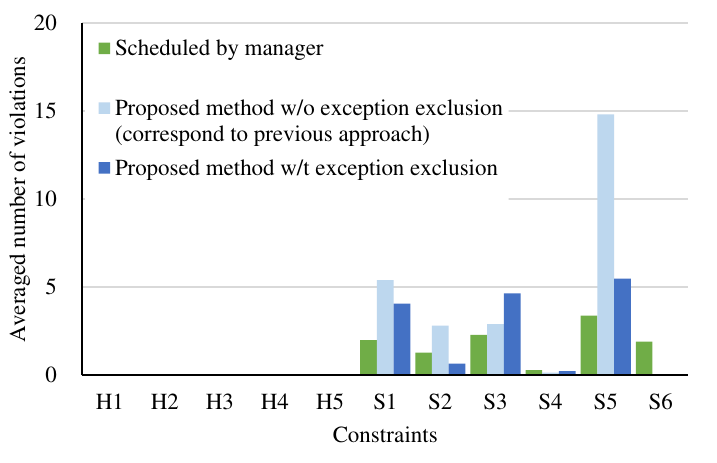}
  \caption{Comparison of the number of violations.}
  \label{fig:vio}
\end{figure}

This experiment compared our proposed method, which removes
exceptional constraints, against a variant that does not.
This latter version (the proposed method without exception exclusion)
was implemented to serve as a baseline corresponding to the
conventional approach such as~\cite{Ben2021,kumar2019automating,paramonov2017tacle}.
Figs.~\ref{fig:ex_created_shift} and \ref{fig:vio} show an example
schedule and the number of violations for hard and soft constraints
obtained through interviews, respectively.
Both cases of the proposed method with and without exception exclusion
using staffing margin and flexibility showed no violations of
any hard constraints, indicating that the proposed method
successfully learned essential constraints.
In addition, exception exclusion reduced the number of violations of
soft constraints 
$S_5$, demonstrating its
effectiveness.

In particular, focusing on the number of violations for $S_6$ in Fig.~\ref{fig:vio}, 
manager-designed schedules often failed to accommodate all staff
requests, especially during months with many requests, averaging two
unsatisfied vacation requests per month.
In contrast, the proposed method, which initially enforced shift
requests as hard constraints, generated conflict-free schedules that
satisfied all requests.

\begin{figure}[t]
\centering
  {\fontsize{4.4pt}{4.4pt}\selectfont
\begin{tabular}{@{}|@{}c@{}|@{}c@{}|@{}c@{}|@{}c@{}|@{}c@{}|@{}c@{}|@{}c@{}|@{}c@{}|@{}c@{}|@{}c@{}|@{}c@{}|@{}c@{}|@{}c@{}|@{}c@{}|@{}c@{}|@{}c@{}|@{}c@{}|@{}c@{}|@{}c@{}|@{}c@{}|@{}c@{}|@{}c@{}|@{}c@{}|@{}c@{}|@{}c@{}|@{}c@{}|@{}c@{}|@{}c@{}|@{}c@{}|@{}c@{}|@{}c@{}|@{}c@{}} \hline
Staff & 1 & 2 & 3 & 4 & 5 & 6 & 7 & 8 & 9 & 10 & 11 & 12 & 13 & 14 & 15 & 16 & 17 & 18 & 19 & 20 & 21 & 22 & 23 & 24 & 25 & 26 & 27 & 28 & 29 & 30 \\\hline
10001 & - & - & - & 1N & 1N & - & D & 1N & 1N & - & D & 1N & 1N & - & D & - & D & 1N & 1N & - & D & 1N & 1N & - & D & D & D & 1N & 1N & D \\\hline
10002 & D & 1N & 1N & - & D & 1N & 1N & - & D & 1N & 1N & - & D & - & D & D & 1N & 1N & - & D & D & - & D & 1N & 1N & - & D & D & - & D \\\hline
10003 & - & D & D & D & - & D & D & D & - & D & D & D & - & D & - & D & D & D & - & D & D & D & - & D & - & - & - & D & D & D \\\hline
10005 & 1N & - & 1N & 1N & - & D & D & D & D & D & 1N & 1N & - & D & 1N & 1N & - & D & D & D & - & D & D & - & - & - & 1N & 1N & - & D \\\hline
10006 & D & - & D & D & - & - & - & D & D & - & - & D & D & D & - & - & - & D & - & - & D & - & D & D & - & - & D & D & D & - \\\hline
10018 & D & D & D & - & D & D & - & - & D & - & - & D & D & - & D & D & - & - & D & - & - & D & - & - & D & D & - & D & - & - \\\hline
10008 & - & D & D & D & D & - & D & D & - & D & D & D & D & - & D & D & D & - & D & D & - & D & D & D & - & D & D & - & D & - \\\hline
10007 & D & D & - & - & D & D & - & - & - & D & - & - & - & D & - & - & D & - & D & - & - & - & - & D & D & D & - & - & D & - \\\hline
10010 & - & - & 2N & 2N & - & - & - & - & - & 2N & 2N & - & 1N & 1N & - & 1N & 1N & - & - & 1N & 1N & - & 1N & 1N & - & 1N & 1N & - & 2N & 2N \\\hline
10011 & 1N & 1N & - & - & - & - & 1N & 1N & - & - & - & - & - & 1N & 1N & - & - & - & 1N & 1N & - & - & - & 2N & 2N & - & - & - & 1N & 1N \\\hline
10050 & - & - & - & - & 1N & 1N & - & - & 1N & 1N & - & - & - & - & - & 2N & 2N & - & - & - & 1N & 1N & - & - & 1N & 1N & - & - & - & 1N \\\hline
10012 & D & - & - & D & D & D & H & D & D & - & D & D & D & - & D & D & - & - & - & D & D & D & - & D & D & D & - & D & D & - \\\hline
10014 & D & 2N & 2N & - & D & D & 2N & 2N & - & D & - & D & 2N & 2N & - & - & D & D & D & D & 2N & 2N & - & D & - & 2N & 2N & - & D & D \\\hline
10013 & D & D & D & - & D & 2N & 2N & - & D & D & - & D & - & D & D & D & 2N & 2N & - & - & D & D & 2N & 2N & - & D & D & D & - & D \\\hline
10015 & - & D & D & D & - & D & D & D & - & D & - & D & D & - & D & D & D & - & D & D & - & D & D & - & D & - & D & D & - & D \\\hline
10044 & - & D & D & - & 2N & 2N & - & D & D & - & D & 2N & 2N & - & - & D & D & D & 2N & 2N & D & - & D & D & 2N & 2N & - & 2N & 2N & - \\\hline
10045 & 2N & 2N & - & 2N & 2N & - & D & D & 2N & 2N & - & D & D & 2N & 2N & - & - & 2N & 2N & - & D & 2N & 2N & - & - & D & D & D & - & D \\\hline
10046 & 2N & - & D & - & - & D & D & 2N & 2N & - & 2N & 2N & - & D & 2N & 2N & - & D & D & 2N & 2N & - & D & - & D & D & 2N & 2N & D & 2N \\\hline
10019 & - & D & - & D & - & D & D & - & D & - & D & - & D & D & - & - & - & D & D & D & - & D & D & D & - & D & D & D & - & D \\\hline
10020 & D & - & - & - & D & - & D & - & D & D & - & - & - & D & D & - & D & - & D & D & - & - & - & D & - & - & D & - & D & - \\\hline
\end{tabular}
}
\caption{An example schedule created using the extracted constraints.}
\label{fig:ex_created_shift}
\end{figure}

\section{Conclusion}
This study proposes a method for extracting facility-specific
constraints from past work schedules to schedule care workers at
long-term care facilities.
This method can extract various constraints by preparing constraint
templates while avoiding extracting exceptional constraints
considering the staffing margin and flexibility, thus
reducing the effort required for
interviews when implementing a scheduling system, which
often impedes the adoption of scheduling systems.
Experiments in a facility with approximately 20 staff members
confirmed that it is possible to extract a set of shifts for each
staff member that can be allocated for a few consecutive days in
addition to primary constraints, such as the types of shifts each staff
member can work and the number of days each shift can be worked.
Importantly,
the proposed method successfully reduced the number of
violations of soft constraints by
excluding exceptional constraints.
In the future, we plan to improve our method so that it can extract
constraints without abstracting the day shifts.

\bibliographystyle{spmpsci}      

\bibliography{papers}

\end{document}